\documentclass[10pt,twocolumn,letterpaper]{article}
\usepackage{cvpr}
\usepackage{times}
\usepackage{epsfig}
\usepackage{graphicx}
\usepackage{amsmath}
\usepackage{amssymb}

\usepackage[breaklinks=true,bookmarks=false]{hyperref}

\cvprfinalcopy 

\def\cvprPaperID{4599} 

\ifcvprfinal\fi
\begin{document}

\title{Convolutional Neural Networks on Randomized Data}

\author{Cristian Ivan\\
	Romanian Institute of Science and Technology\\
	Cire\c{s}ilor 29, 400487 Cluj-Napoca, Romania\\
	{\tt\small ivan@rist.ro}
}

\maketitle

\begin{abstract}
	Convolutional Neural Networks (CNNs) are build specifically for computer vision 
	tasks for which it is known that the input data is a hierarchical structure 
	based on locally correlated elements. The question that naturally arises is what 
	happens with the performance of CNNs if one of the basic properties of the data is 
	removed, e.g. what happens if the image pixels are randomly permuted? Intuitively 
	one expects that the convolutional network performs poorly in these circumstances 
	in contrast to a multilayer perceptron (MLPs) whose classification accuracy should 
	not be affected by the pixel randomization. 
	This work shows that by randomizing image pixels the hierarchical structure of the
	data is destroyed and long range correlations are introduced which standard CNNs 
	are not able to capture. We show that their classification accuracy is heavily 
	dependent on the class similarities as well as the pixel randomization process.
	We also indicate that dilated convolutions are able to recover some of the pixel 
	correlations and improve the performance.

\end{abstract}

\section{Introduction}\label{introduction}

Convolutional Neural Networks are inspired by the visual system of living 
organisms and are built to exploit the 2D structure of natural images \cite{lecun-98}. 
The receptive field offered by the convolution kernels greatly reduces the number
of trainable parameters and increases the performance of these networks as
compared to fully connected feed forward networks. CNNs are used not only for
visual tasks but also on other kind of data where local correlations are still
present.

One question that can be asked is what is the performance of CNNs 
when trained on images where the individual pixels are randomly permuted? 
Since the local spatial structure of an image is destroyed one would expect that 
a convolutional network is not able to find representative features 
of the data and the accuracy should be very low.

For this study two types of feed-forward networks are used for image 
classification tasks. CNNs and MLPs are trained on natural images and their 
pixel-wise permutations. The hyper-parameters of the networks are kept the same
throughout the performed experiments and are trained for the same number of 
epochs. The MLP is used as a baseline and sanity check for the analysis and, 
due to the known limitations it has when trained on complex image databases, 
its performance is not intended to
be used as a direct comparison to the CNN performance.

The chosen architecture of the CNN follows closely the VGG16 network 
\cite{DBLP:journals/corr/SimonyanZ14a} and the MLP consists of the last 
fully connected layers of the CNN. The only difference between the CNN 
used in this paper and the VGGNet is the total number of parameters 
since they are trained on much smaller images, 28 and 32 pixels per side. 
Both networks in this study
use \textit{adam} optimizer \cite{DBLP:journals/corr/KingmaB14} with a 
learning rate of 0.0001 and a decay of $10^{-6}$.
The training procedure does not use any form of data augmentation.

We perform three types of experiments where we compare the classification
accuracy of a CNN trained on natural images with the accuracy of a CNN with
the same architecture trained on images whose contents are randomized based 
on three different procedures.

In the first experiment the order of the image pixels is fully shuffled. 
The classification accuracy of the CNN is investigated for increasingly complex 
datasets: MNIST \cite{mnist}, Fashion-MNIST \cite{fashionmnist} and CIFAR10 \cite{cifar10}. In the second and third experiment we develop two different parametrized 
methods for controlling the image randomization and investigate the CNN 
classification accuracy on only the CIFAR10 dataset.

\section{Pixel-wise permutations}\label{pixelwisepermutations}

If we consider the pixels of an $(n \times n)$ image in row-major order as the set
$\left\lbrace 1,2,3,4,5, \cdots, n^{2} \right\rbrace $ then a pixel-wise permutation can be 
expressed in Cauchy's two-line notation as: 
\begin{equation}\label{permutation}
\sigma= \Bigl(\begin{matrix}
1 & 2 & 3 & 4  		&5	&  \cdots & 	n^{2}\\
4 & 5 & 1 & n^{2} 	&3	&  \cdots & 	2 
\end{matrix}\Bigr)
\end{equation}\label{permutationeq}

where the second line represents the new arrangement of the original pixels in the
permuted image. Therefore pixel 1 from the original image is moved to position 4
in the permuted image, pixel 2 is moved to position 5 etc.

\subsection{MNIST}

\begin{figure}[h]
	\begin{center}
		\includegraphics[width=1\linewidth]{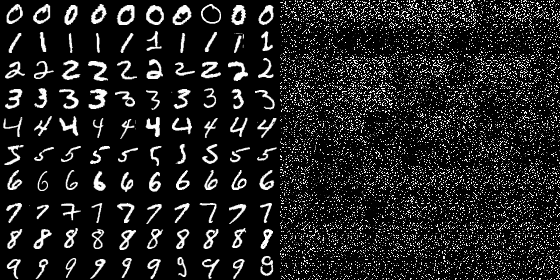}
	\end{center}
	\caption{Left panel: random samples of MNIST natural images; 
		right panel:  pixel-wise randomization of the same sample}
	\label{fig:trainimages-mnist}
\end{figure}

A sample of both natural and pixel-wise permuted images is shown in Figure
\ref{fig:trainimages-mnist}: the left panel shows handwritten digits 
from the MNIST dataset  and the right panel shows a pixel permutation
of those images. The same permutation is applied on all train and test images.

\begin{figure}[h]
	\begin{center}
		\includegraphics[width=1\linewidth]{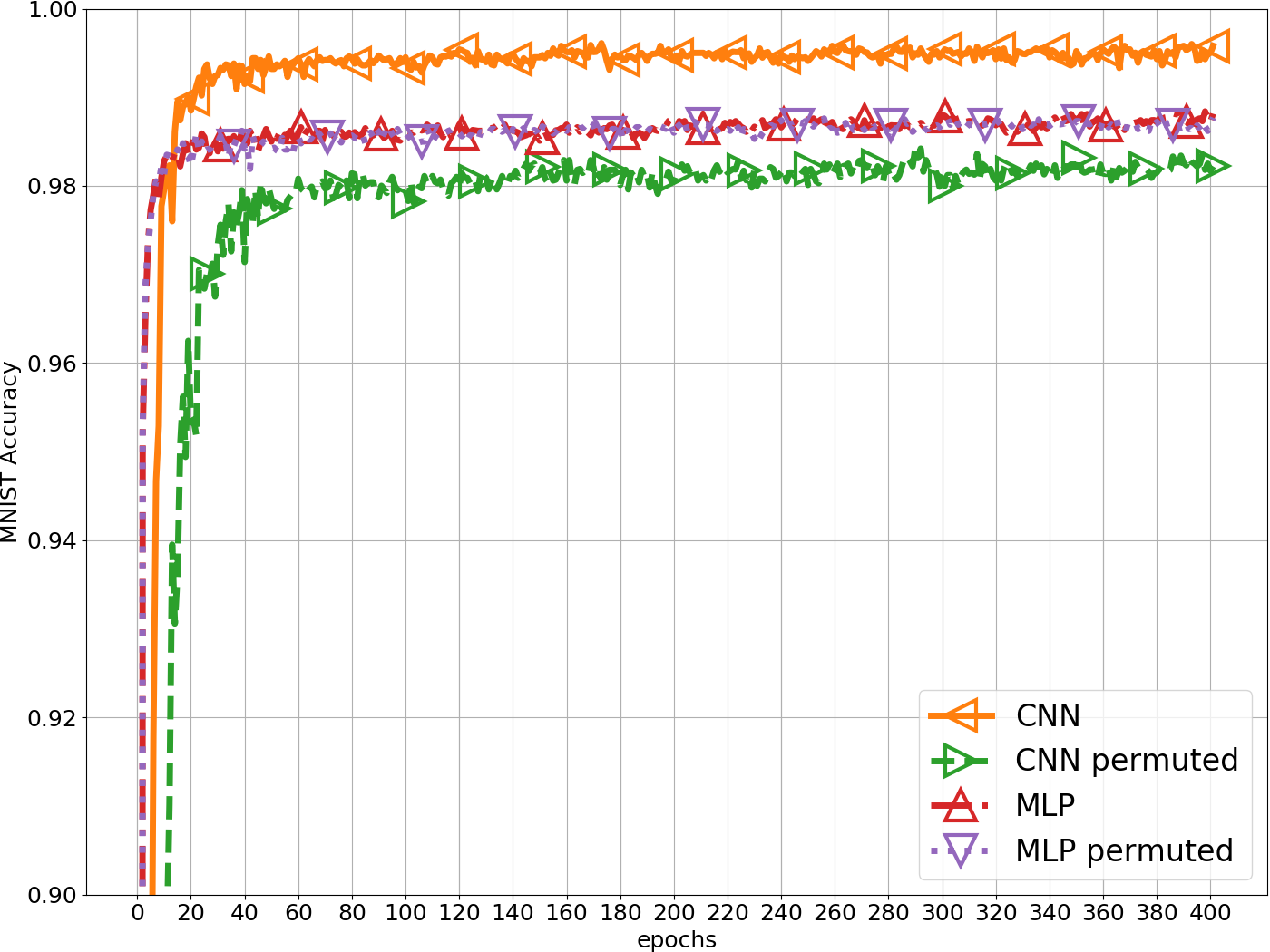}
	\end{center}
	\caption{Accuracy of a CNN and MLP running on MNIST images and their 
		permutations (color online)}
	\label{fig:perfMNIST}
\end{figure}

The test data accuracy of the networks can be seen in Figure \ref{fig:perfMNIST}.
The CNN trained on natural images reaches an accuracy of $99.5\%$
while the one trained on permuted images shows a delayed learning curve 
as well as a consistently lower performance.

The performance of the MLP trained on both natural and pixel-wise permuted 
images is almost identical throughout the entire training phase. 
A more interesting observation is that the MLP, which consists of
only the last fully connected layers of the CNN, has higher performance 
on permuted images than the CNN at every point during the training phase.

\subsection{Fashion-MNIST}

\begin{figure}[h]
	\begin{center}
		\includegraphics[width=1\linewidth]{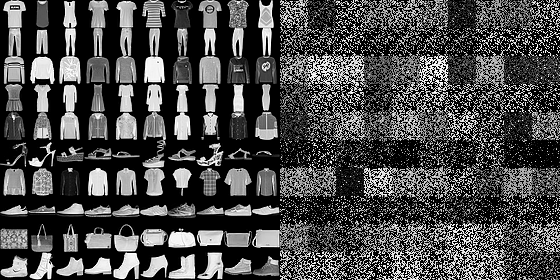}
	\end{center}
	\caption{Left panel: random sample of Fashion-MNIST images; 
		right panel: pixel-wise randomization of the same sample}
	\label{fig:trainimages-fashion}
\end{figure}

The Fashion-MNIST dataset 
has a higher complexity than MNIST and poses a greater difficulty for the 
networks. The left panel of Figure \ref{fig:trainimages-fashion} 
shows a sample of the greater variability
of clothing images and their pixel-wise permutations.

\begin{figure}[h]
	\begin{center}
		\includegraphics[width=1\linewidth]{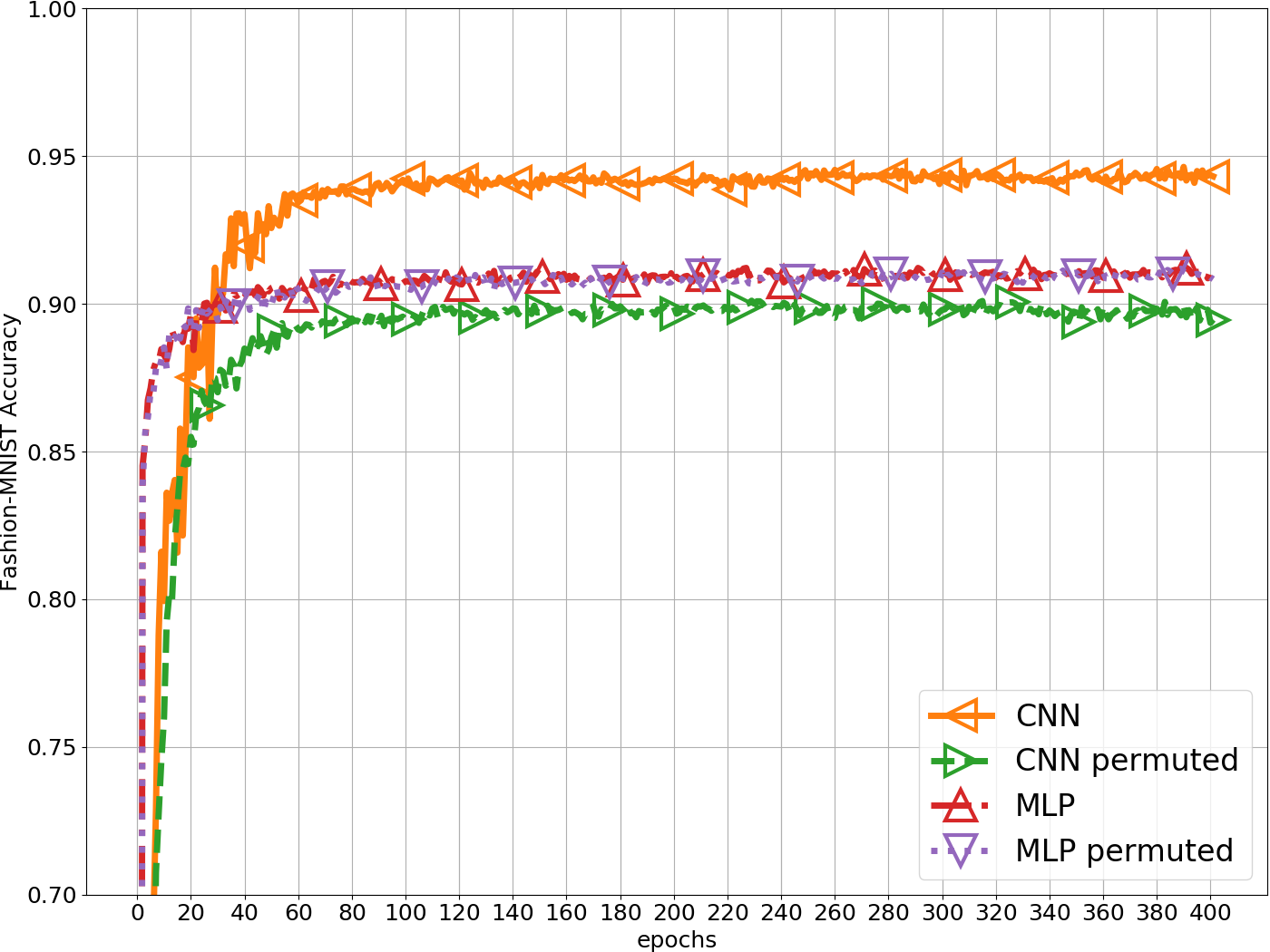}
	\end{center}
	\caption{Accuracy of a CNN and MLP trained on Fashion-MNIST dataset.}
	\label{fig:perfFashion}
\end{figure}

Figure \ref{fig:perfFashion} shows a lower performance, as compared to MNIST,
on natural and pixel-wise randomized images for both types of networks. 
The peak performance of the CNN trained on the natural images is $94.3\%$, decreasing to
$89.6\%$ when trained on the randomized data set.
The same behaviour is observed as in the previous experiment: the test accuracy 
of the CNN trained on permuted images is consistently below the MLP.

\subsection{CIFAR10}

\begin{figure}[h]
	\begin{center}
		\includegraphics[width=1\linewidth]{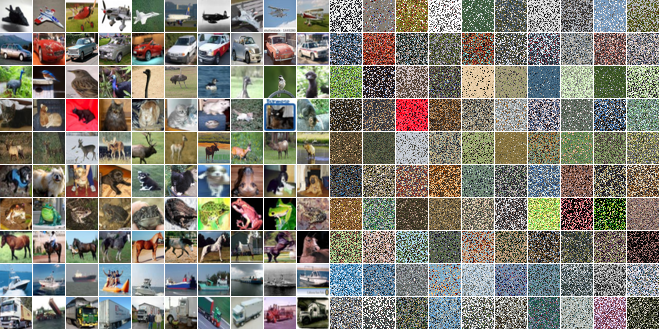}
	\end{center}
	\caption{Left panel: random sample of CIFAR10 images; 
		right panel: pixel-wise randomization of the same sample.}
	\label{fig:trainimages-CIFAR10}
\end{figure}

The CIFAR10 dataset is far more complex than the previous two: the images 
are in 3 color channels, the objects are not centered, they are less 
similar to each other and the background is not uniform. 
Figure \ref{fig:trainimages-CIFAR10} shows the natural and pixel-wise 
randomization images, for the latter the identical randomization procedure 
being applied to all channels. The various labels are listed in table 
\ref{tab:cifar10labels}.

Training a CNN on this database reveals an even lower classification accuracy 
when using natural images, slightly below $90\%$, and a dramatic performance decrease 
when training on pixel-wise permuted images, reaching only about $57\%$, while 
the MLP is invariant under this type of transformation.

\begin{figure}[h]
	\begin{center}
		\includegraphics[width=1\linewidth]{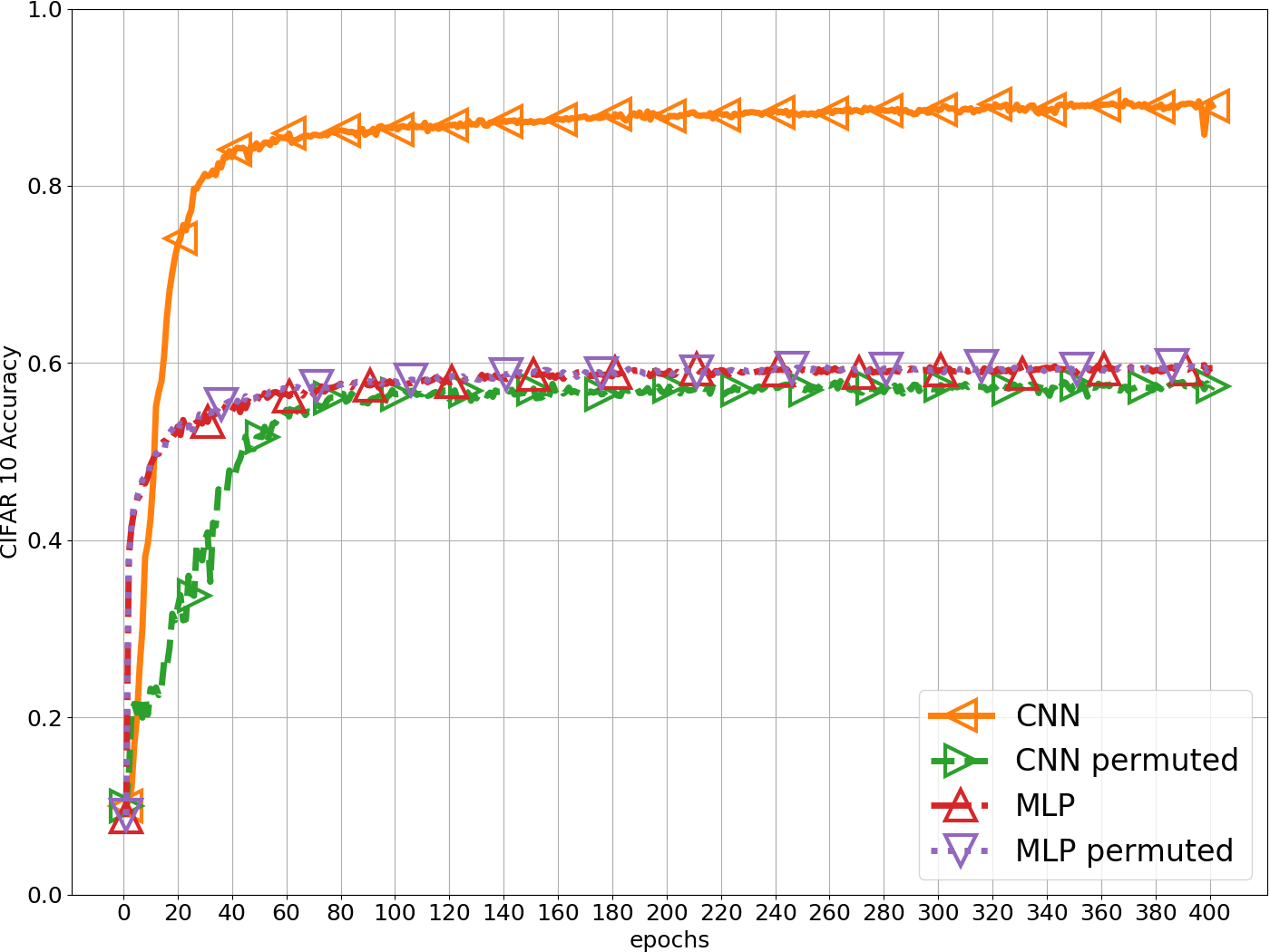}
	\end{center}
	\caption{Accuracy of a CNN and MLP running on CIFAR10 images and their permutations.}
	\label{fig:perfCIFAR10}
\end{figure}

Figure \ref{fig:perfCIFAR10} shows the evolution of the accuracy on the test 
images.
There is again the trend of the accuracy of 
the CNN trained on permuted images to stay consistently below the baseline 
performance of the MLP. Table \ref{tab:accuracyjumps} 
summarizes the peak accuracies of the networks trained on all three datasets.

\begin{table}[h]
	\begin{center}
		
		\begin{tabular}{|c|c|c|c|c|}
			\hline
			&  \multicolumn{2}{|c|}{CNN} & \multicolumn{2}{|c|}{MLP}\\ \hline
			\textbf{Images} 	& natural  	& permuted 	  & natural  	& permuted 	\\ \hline
			\textbf{MNIST}      & 99.5\% 	& 98.2\%      & 98.7\% 		& 98.6\%  	\\ \hline
			\textbf{Fashion}    & 94.3\% 	& 89.6\%      & 91.0\% 		& 90.9\%  	\\ \hline
			\textbf{CIFAR10}    & 88.9\% 	& 57.3\%      & 59.3\% 		& 59.3\%	\\ \hline
		\end{tabular}	
	\end{center}
	\caption{Accuracy of a CNN and MLP trained on natural images and their permutations.}
	\label{tab:accuracyjumps}
\end{table}

\section{Image patch permutations}\label{imagepatchpermutations}

\begin{figure}[h]
	\begin{center}
		\includegraphics[width=0.90\linewidth]{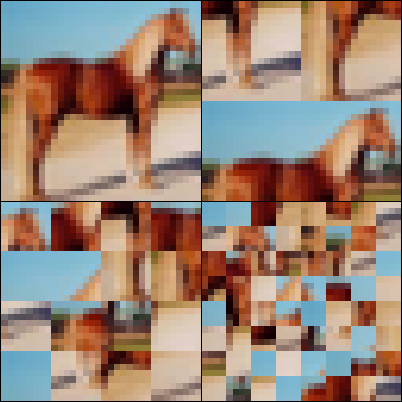}
	\end{center}
	\caption{Example of a natural image (top-left) and its patch-wise randomization of size 16, 8 and 4.}
	\label{fig:datapermutedpatches4816cifar10}
\end{figure}

To further investigate the behaviour of the network on data randomization a
parametrized method is developed in order to gain a better control on the 
randomization process.
The images are sliced in square patches which are then shuffled in the same
manner as described by equation \ref{permutationeq}, where the numbers, instead
of image pixels, denote the image patches.
The parameter that controls the randomization is the size of the 
patch: a size of 1 is equivalent to a full pixel-wise permutation and a size of 
32 is equivalent with the natural image. Figure \ref{fig:datapermutedpatches4816cifar10} 
shows a few examples of permutations with intermediate patch sizes.

\begin{figure}[t]
	\begin{center}
		\includegraphics[width=1\linewidth]{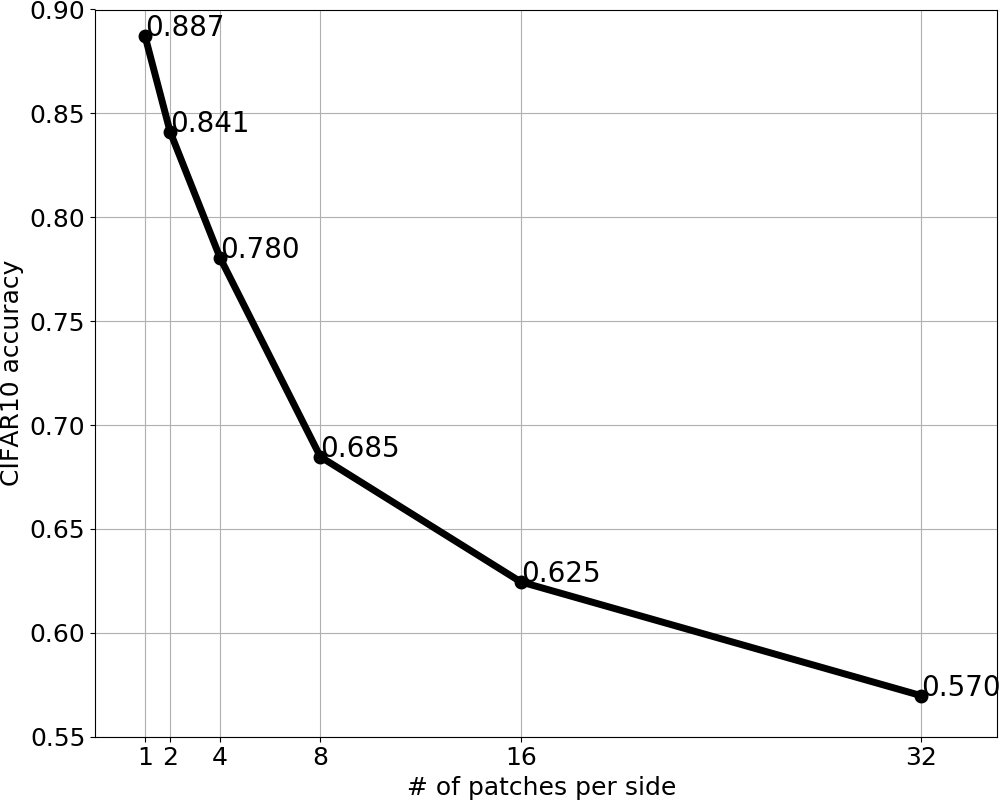}		
	\end{center}
	\caption{CNN classification accuracy as a function of the number of patches 
		used for randomization. 32 means the image was sliced in 32 patches of $[1\times1]$ pixels, 16 means an image slices in 16 patches of $[2\times2]$ pixels etc.}
	\label{fig:AccRandomPatchDependence}
\end{figure}

Alternatively one can consider the 
randomization parameter the number of slices the image is cut into along an axis
which results in a convenient series of powers of 2.
The classification accuracy of the CNN trained with this parametrization is 
displayed in Figure \ref{fig:AccRandomPatchDependence}. It shows how
strongly the CNN performance is influenced by the size of the patches used for the
randomization. Even when cutting the images in 4 slices per side the CNN looses in 
accuracy considerably.

\section{Local permutations}\label{localpermutations}

In this experiment the image pixels are randomized inside a restricted 
neighborhood. The algorithm is as follows: the image is scanned in a row-major 
order and each pixel is swapped with a randomly chosen pixel inside a neighborhood of $D$ pixels. \begin{figure}[b]
	\begin{center}
		\includegraphics[width=0.9\linewidth]{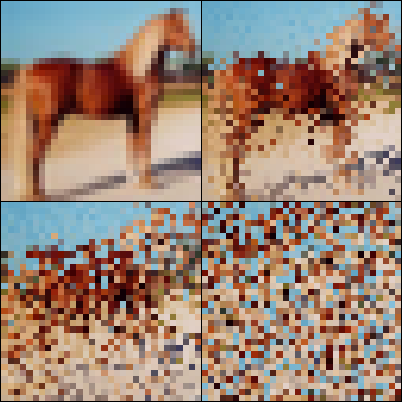}
	\end{center}
	\caption{Examples of a natural image and its pixel-wise randomization of distances 1, 4 and 16.}
	\label{fig:trainimagesdistancerandomization}
\end{figure}
A zero distance is equivalent to no permutation and a 32 distance
to a full pixel-wise permutation. There are no restrictions on the number of times
a pixel can be moved. Hence there is a non-zero chance that a pixel might migrate 
a distance longer than $D$. This effect is rather small, as can be seen in 
Figure \ref{fig:trainimagesdistancerandomization} where the top-right image
shows a local permutation with distance 1.

\begin{figure}[t]
	\begin{center}
		\includegraphics[width=1\linewidth]{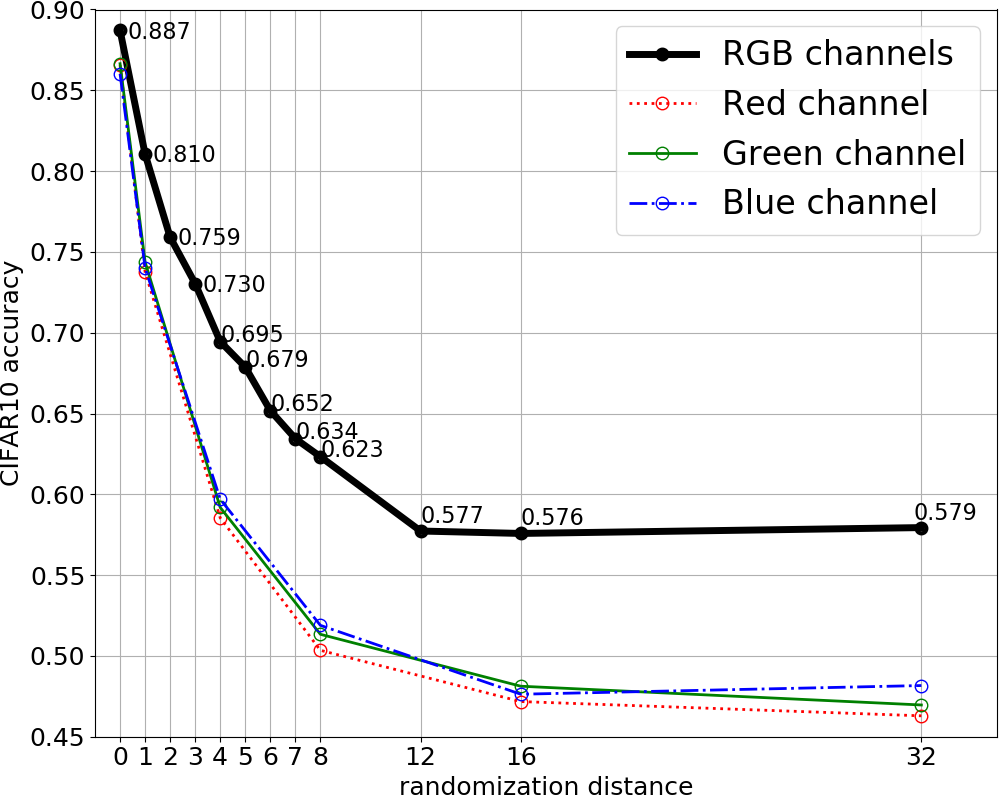}
	\end{center}
	\caption{CNN classification accuracy dependence on the randomization distance. The black curve correspond to a training on the RGB channels of CIFAR10 images and the red, green and blue curves show the accuracy for the corresponding color channel.}
	\label{fig:AccRandomLocalDependence}
\end{figure}

Figure \ref{fig:AccRandomLocalDependence} shows the performance of the CNN as a
function of several distances, ranging from 0 up to the size of the whole image.
The larger the randomization distance is the less features the $[3\times3]$
kernels are able to capture and the accuracy quickly decreases, reaching the same 
level as a CNN trained on fully pixel-wise randomized images.

It is interesting to note how the CNN classification accuracy decreases when 
trained on separate image channels. Humans are able to recognize objects in 
images being either in color or gray-scale, with shape being the crucial factor 
in the classification process. The CNN correlates not only local pixels in 
terms of shape but also local pixels in terms of colors. 
Training on natural images and single channels results in a $2\%$ performance 
decrease as compared to training on all channels. When the local pixel 
correlations are destroyed the network performance degrades considerably more 
when trained on just single color channels.
Comparing the color curves in Figure \ref{fig:AccRandomLocalDependence} with 
the black curve indicates that the network relies also on color correlations when
doing classification, since the same randomization procedure is applied on all
channels.

The color curves also show a slight data bias due to the growing discrepancy 
between the classification accuracy on the three different channels. Training on
just the red channel results in the worst performance for all randomization 
distances indicating that there might be less information in this channel.

\section{Discussion}\label{discussion}

Figures \ref{fig:perfMNIST}, \ref{fig:perfFashion} and \ref{fig:perfCIFAR10} 
show common and consistent behaviors: the performance of the CNN drops 
when switching from training on natural to permuted images and always stays below the 
accuracy of the MLP throughout the whole training phase. The accuracy decrease 
cannot be attributed to a particular configuration of the network's initial 
random weights because repeated trainings display the same effect. 

In Figures \ref{fig:AccRandomPatchDependence} and \ref{fig:AccRandomLocalDependence} 
we have shown how the performance of the CNN changes as the image pixels are gradually
randomized. The stronger the randomization is the lower 
the accuracy of the network becomes. By applying the same permutation 
to all examples the local patterns that repeat within the same image (e.g. edges at
various inclinations) are destroyed but 
the intra-class similarities and inter-class differences are kept.
The accuracy of the CNN decreases because the convolution kernels can not
find hierarchical features in the randomized images but it does
not decrease to the level of random guessing since the separation between
classes still remains. It is unclear how to quantify the accuracy decrease 
the randomization induces.

The convolutional network has a built-in infinitely strong prior which 
constrains the values of some parameters to zero making it highly sensitive 
to the spatial structure of the data \cite{Goodfellow-et-al-2016}. The more
the local pixel correlation is removed the lower the classification
accuracy becomes.
The performance of a CNN has at least two independent components:
\begin{enumerate}
	\item like any other feed forward ANN, the network finds intra-class 
	similarities and inter-class difference via gradient descent.
	\item the kernels learn local patterns occurring repeatedly 
	within different regions of the \textbf{same image}, within different 
	examples of the \textbf{same class} and within different examples of \textbf{different classes}.	
\end{enumerate}
Natural images are structures where local correlations are important.
This is the reason why edges, corners, Gabor-like features etc. 
area learned by the CNN's first few layers when trained on natural 
images \cite{erhan2009visualizing}.
Combining these basic building blocks in complex hierarchical structures 
results in large varieties of images and objects. At a more profound 
level it is speculated that this is the reason why deep learning works so 
well in practice and that the very structure of the universe is basically 
a large hierarchy of simple elements \cite{Lin2017}.
If pixels are moved in random positions across the image domain their initial 
local correlation is destroyed an it becomes a long range correlation. If there
is a correlation among pixels from more distant locations a kernel spread on
a wider area of the image is more likely to capture them than a local 
$[3\times3]$ kernel.

\begin{figure}[t]
	\begin{center}
		\includegraphics[width=1\linewidth]{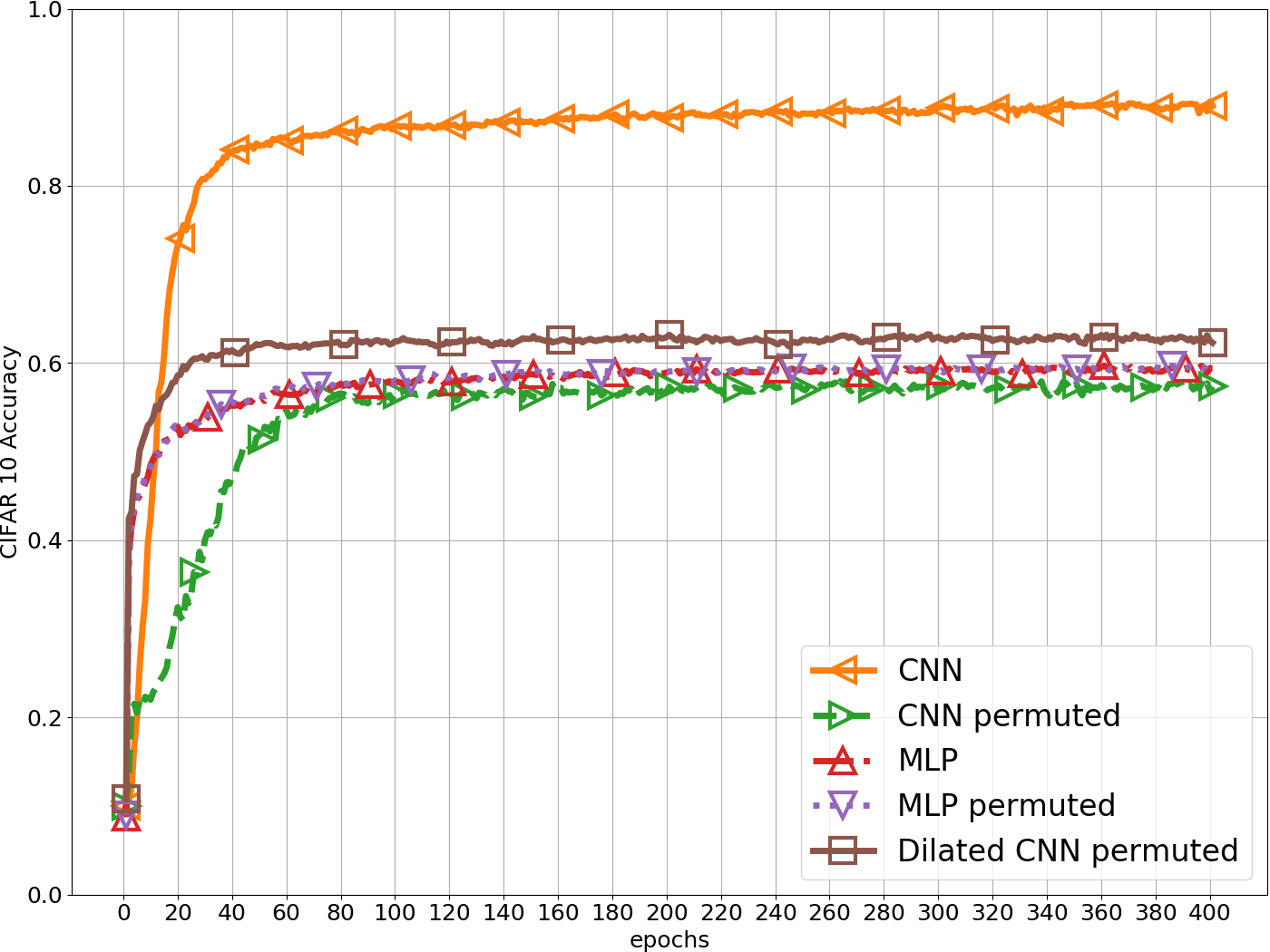}
	\end{center}
	\caption{Classification accuracy of the VGG16-like CNN, MLP and CNN with dilated
		convolutions trained on natural and permuted images.}
	\label{fig:perfDilated}
\end{figure}

Based on this idea we experimented with a modified version of the CNN where
we replaced the whole stack of many convolution layers with two identical 
dilated convolutional layers: 64 filters with $[4\times4]$ kernels, dilation 
rate of 4 and a stride of 1. 
Figure \ref{fig:perfDilated} shows a $62\%$ accuracy when training on pixel-wise
permuted images indicating that some of the long range correlations are captured 
by the dilated kernels. There is a problematic aspect of this approach in that 
it does not permit arbitrary number of such layers due to the inevitable image 
size reduction of this convolution operation. We have also tested a single
convolutional layer with an $[8\times8]$ kernel with a stride of 4 but it did 
not exceed $60\%$ accuracy. Stacking many fully connected layers together did 
not surpass the classification accuracy of the dilated convolutional network
indicating that deep MLPs are not a solution for this type of images.

\section{Data properties}\label{dataproperties}

In this section we will present some basic properties of the data and
show how strong the correlation between training on natural images vs. 
pixel-wise permutations is.  This, in turn, indicates the underlying 
structure of the data even when individual image pixels are randomly 
permuted inside the images.

\subsection{Fashion-MNIST}

\begin{figure}[h]
	\begin{center}
		\includegraphics[width=1\linewidth]{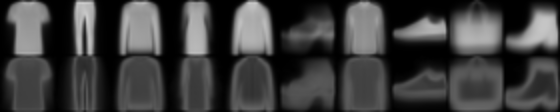}
	\end{center}
	\caption{Mean (upper row) and standard deviation (lower row) of Fashion-MNIST training images; the corresponding labels are listed in Table \ref{tab:fashionlabels}.}
	\label{fig:meansstdsfashion}
\end{figure}

The Fashion-MNIST dataset is more complex than MNIST, as can be seen from
the classification accuracy in Table \ref{tab:accuracyjumps}. But the
greater difficulty of this dataset comes not from the objects alone, but
rather from  the strong similarities between differently labeled 
classes. The left panel of Figure \ref{fig:trainimages-fashion} shows 
how similar many examples of \textbf{shirt} look like \textbf{T-shirt} 
or \textbf{coat}. The average Fashion-MNIST images are shown in the top 
row of Figure \ref{fig:meansstdsfashion} and the standard deviations in the
bottom row.

\begin{table}[h]
	\begin{center}
		\begin{tabular}{|c|c|c|c|}
			\hline 
			\textbf{0} & T-shirt/top & \textbf{5} & Sandal \\ 
			\hline 
			\textbf{1} & Trouser &\textbf{6} & Shirt \\ 
			\hline 
			\textbf{2} & Pullover & \textbf{7} & Sneaker \\ 
			\hline 
			\textbf{3} & Dress & \textbf{8} & Bag \\ 
			\hline 
			\textbf{4} & Coat & \textbf{9} & Ankle boot \\ 
			\hline 
		\end{tabular} 
	\end{center}
	\caption{Fashion-MNIST labels}
	\label{tab:fashionlabels}
\end{table}

\begin{figure}
	\begin{center}
		\includegraphics[width=1\linewidth]{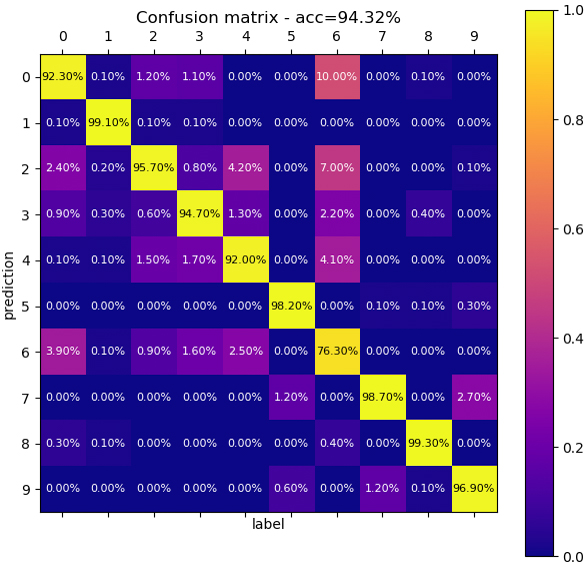}
	\end{center}
	\caption{Confusion matrix for CNN on natural Fashion-MNIST images.}
	\label{fig:cmcnnsfashion}
\end{figure}

The confusion matrix in Figure \ref{fig:cmcnnsfashion} illustrates how 
often label 6 (\textbf{shirt}) is misclassified as \textbf{T-shirt},  \textbf{coat}, \textbf{pullover} and \textbf{dress}. If it were not for these strong 
similarities between labels 0, 2, 4 and 6 the overall performance of the CNN 
would be higher, most of the other categories having a classification accuracy significantly more than 90\%. Other similar correlations can be seen in the non-zero entries in the $7^{th}$
column, which indicate that the network identified shoe-like features in 
\textbf{sandal}, \textbf{sneaker} and \textbf{ankle boots}.

\subsection{CIFAR10}

CIFAR10 has considerably more variation than Fashion-MNIST and the mean
and standard deviation figures are not relevant for this kind of analysis
as the distributions are much more uniform. The image features and class
commonalities are difficult to see by the naked eye and more sophisticated
statistical tools are needed to shed light on the intra-class and
inter-class correlations.
However, by investigating the confusion matrix one can identify the
correlations the networks learn.

\begin{figure}[b]
	\begin{center}
		\includegraphics[width=1\linewidth]{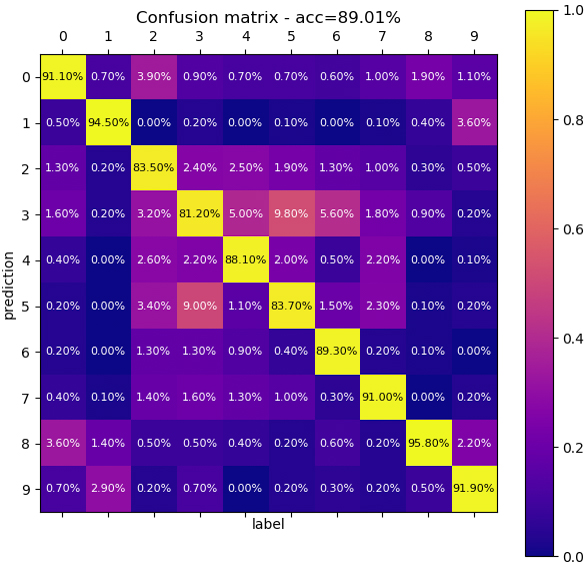}
		\includegraphics[width=1\linewidth]{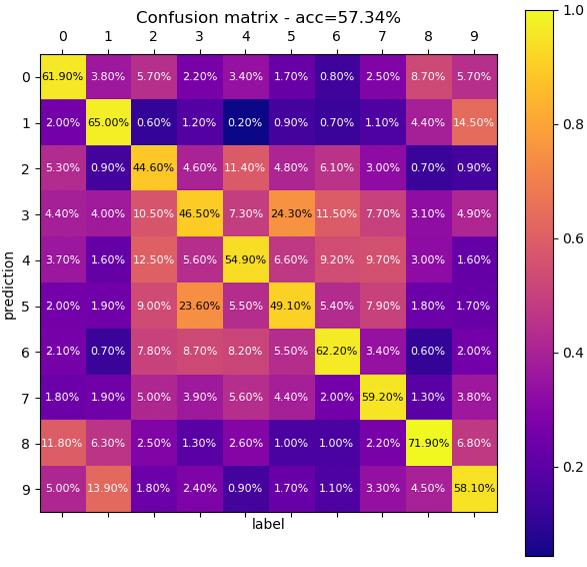}
	\end{center}
	\caption{Confusion matrix for CNN on natural CIFAR10 images (top panel) and pixel-wise randomizations (bottom).}
	\label{fig:cmcnnscifar10}
\end{figure}

Figure \ref{fig:cmcnnscifar10} 
shows the confusion matrices of the CNN trained on natural (top-panel) and pixel-wise permutations, respectively (bottom-panel). 
For training on natural images the top two highest accurate predictions
are for the \textbf{ship} and \textbf{automobile}. The largest classification error is made in the 
case of the \textbf{cat} which is $9\%$ of the time misclassified 
as a \textbf{dog}. 
Notice that the confusion matrices are only approximately symmetric and the 
reverse misclassifications are slightly different, e.g.\ \textbf{dog} confused as a \textbf{cat} 
is about $9.8\%$.

\begin{figure}[b]
	\begin{center}
		\includegraphics[width=1\linewidth]{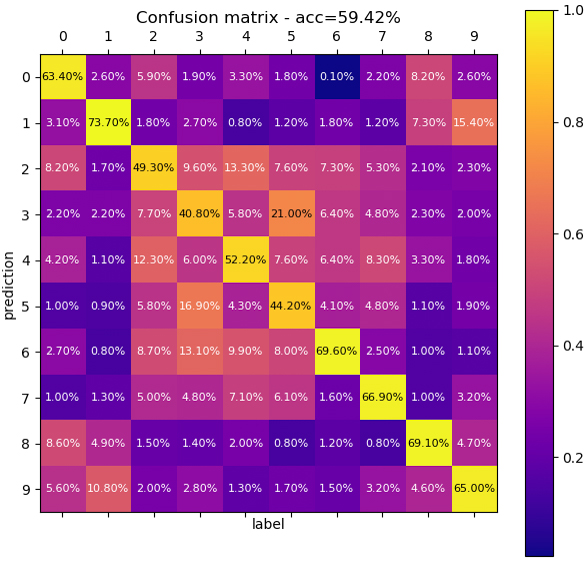}
		\includegraphics[width=1\linewidth]{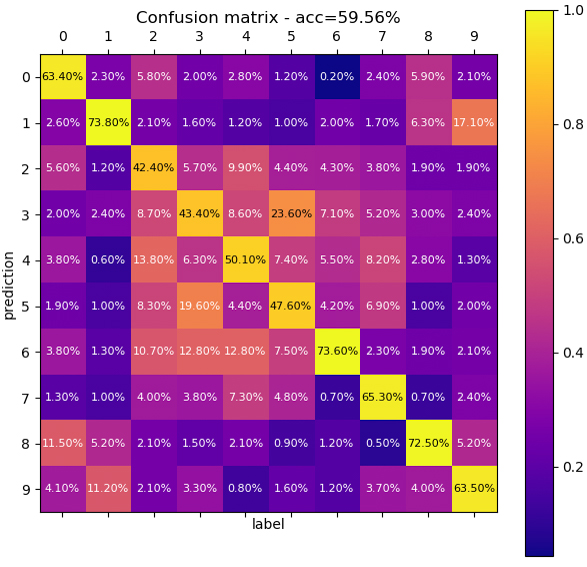}
	\end{center}
	\caption{Confusion matrix for MLP on natural CIFAR10 images (top) and pixel-wise permutation (bottom).}
	\label{fig:cmmlpscifar10}
\end{figure}

\begin{table}[t]
	\begin{center}
		\begin{tabular}{|c|c|c|c|}
			\hline 
			\textbf{0} & airplane & \textbf{5} & dog \\ 
			\hline 
			\textbf{1} & automobile &\textbf{6} & frog \\ 
			\hline 
			\textbf{2} & bird & \textbf{7} & horse \\ 
			\hline 
			\textbf{3} & cat & \textbf{8} & ship \\ 
			\hline 
			\textbf{4} & deer & \textbf{9} & truck \\ 
			\hline 
		\end{tabular} 
	\end{center}
	\caption{CIFAR10 labels}
	\label{tab:cifar10labels}
\end{table}

Although overall the classification accuracy drops from $89\%$ to $57\%$ when switching from training on natural images to permuted images the correlations 
made by the network in one case are very similar to the other case. 
One can observe the same central pattern in both figures. 
The top two classification accuracies are the same - \textbf{ship} and \textbf{automobile} and the \textbf{cat-dog} misclassification remains 
still the highest.

Figure \ref{fig:cmmlpscifar10} shows the confusion matrices of the MLP trained
on natural and pixel-wise permuted images. Unlike the CNN, the overall accuracy difference is very small, $0.14\%$, but some of the individual accuracies show 
a relatively high variation, the largest deviation being for \textbf{bird} and
\textbf{frog} with $-6.9\%$ and $+4\%$, respectively. A similarly strong 
\textbf{cat-dog} confusion is done also for this network.

We can obtain a quantification of the
similarities between the two CNNs and MLPs by calculating the 
Pearson correlation coefficient 
between the prediction a network does for a particular category when trained 
on natural images and the prediction it makes for the same category when 
training on permuted images. In other words for the CNN we correlate the upper 
columns of Figure \ref{fig:cmcnnscifar10} with the lower columns from the same 
figure. The same calculations are performed for the MLP confusion matrices in 
Figure \ref{fig:cmmlpscifar10}.
Thus we obtain 10 correlation coefficients which are summarized in table 
\ref{tab:cmcorrelations} together with the average correlation coefficient.

\begin{table}[h]
	\begin{center}
		\begin{tabular}{|c|c|c|}
			\hline 
			\textbf{Class}& \textbf{CNN} & \textbf{MLP}  \\ 
			\hline 
			\hline
			airplane& 0.951 & 0.862 \\ 
			\hline 
			automobile& 0.974 & 0.995 \\ 
			\hline 
			bird& 0.745 & 0.974 \\ 
			\hline 
			cat& 0.963 & 0.960 \\ 
			\hline 
			deer& 0.645 & 0.902 \\ 
			\hline 
			dog& 0.982 & 0.983 \\ 
			\hline 
			frog& 0.753 & 0.916 \\ 
			\hline 
			horse& 0.907 & 0.926 \\ 
			\hline 
			ship& 0.857 & 0.967 \\ 
			\hline 
			truck& 0.925 & 0.992 \\ 
			\hline 
			\hline
			\textbf{Mean}& \textbf{0.870} &\textbf{0.947}  \\ 
			\hline 
		\end{tabular} 
	\end{center}
	\caption{Correlations between the predictions of networks trained on natural images and pixel-wise permutations of CIFAR10 images.}
	\label{tab:cmcorrelations}
\end{table}

\section{Conclusions and further studies}

This paper presents the limitations of convolutional neural networks when
trained on images where individual pixel positions have been randomly permuted.
We show a comparison of classification accuracies between CNNs trained on
natural images and pixel-wise permutations together with the performance 
of an MLP as baseline. The absolute value of the accuracies are not relevant
to the study, just the relative performances.

We have shown that the use of standard convolutional networks is inappropriate 
for cases where image pixels are permuted inside the image domain. We create 
long range correlations which can be better captured by kernels covering
larger image areas than standard localized kernels.
We have shown that by applying the same permutation to all images from the
dataset there is still an underlying structure which can be discovered by 
neural networks. This suggests the possibility for further model improvements.
It is important to design networks with architectures that can be
invariant or at least less sensitive to data permutations or other types
of data encryption. There are studies 
\cite{DBLP:journals/corr/abs-1711-05189} 
which show that CNNs are still capable of classifying encrypted images, although
in that particular case the transformation is a homomorphic encryption which 
preserves more of the data structure than the pixel randomizations do. 
This further raises the question whether it is possible to train a 
network on encrypted data but then be able to reconstruct the initial 
data once a few examples of human interpretable data become available.

Many types of analyses, where data is not necessary locally correlated,
would benefit from such empowered architectures. For example, high energy physics experiments
require the analysis of large data sets from particle collisions
where the data appears on an event-by-event basis as random tracks in the 
detectors. However there are very strong underlying correlations since the subatomic processes obey the laws of physics. Often, \cite{Duarte:2018ite} 
\cite{DBLP:journals/corr/abs-1708-07034}, CNNs are used for the
analysis of features identified by physicists through standard methodologies. 
Networks which would perform well on seemingly random data would be of great
use for this kind of studies.

Other domains could also greatly benefit from more powerful networks 
designed specifically for capturing long range correlations, situations which 
can easily arise in the experimental physical sciences.


\begin{thebibliography}{10}\itemsep=-1pt
	
	\bibitem{Duarte:2018ite}
	J.~Duarte et~al.
	\newblock {Fast inference of deep neural networks in FPGAs for particle
		physics}.
	\newblock {\em JINST}, 13(07):P07027, 2018.
	
	\bibitem{erhan2009visualizing}
	D.~Erhan, Y.~Bengio, A.~Courville, and P.~Vincent.
	\newblock Visualizing higher-layer features of a deep network.
	\newblock {\em University of Montreal}, 1341(3):1, 2009.
	
	\bibitem{Goodfellow-et-al-2016}
	I.~Goodfellow, Y.~Bengio, and A.~Courville.
	\newblock {\em Deep Learning}.
	\newblock MIT Press, 2016.
	\newblock \url{http://www.deeplearningbook.org}.
	
	\bibitem{DBLP:journals/corr/abs-1711-05189}
	E.~Hesamifard, H.~Takabi, and M.~Ghasemi.
	\newblock Cryptodl: Deep neural networks over encrypted data.
	\newblock {\em CoRR}, abs/1711.05189, 2017.
	
	\bibitem{DBLP:journals/corr/KingmaB14}
	D.~P. Kingma and J.~Ba.
	\newblock Adam: {A} method for stochastic optimization.
	\newblock {\em CoRR}, abs/1412.6980, 2014.
	
	\bibitem{cifar10}
	A.~Krizhevsky.
	\newblock Learning multiple layers of features from tiny images.
	\newblock 2009.
	
	\bibitem{mnist}
	Y.~LeCun and C.~Cortes.
	\newblock {MNIST} handwritten digit database.
	\newblock 2010.
	
	\bibitem{Lin2017}
	H.~W. Lin, M.~Tegmark, and D.~Rolnick.
	\newblock Why does deep and cheap learning work so well?
	\newblock {\em Journal of Statistical Physics}, 168(6):1223--1247, Sep 2017.
	
	\bibitem{DBLP:journals/corr/abs-1708-07034}
	C.~F. Madrazo, I.~H. Cacha, L.~L. Iglesias, and J.~M. de~Lucas.
	\newblock Application of a convolutional neural network for image
	classification to the analysis of collisions in high energy physics.
	\newblock {\em CoRR}, abs/1708.07034, 2017.
	
	\bibitem{DBLP:journals/corr/SimonyanZ14a}
	K.~Simonyan and A.~Zisserman.
	\newblock Very deep convolutional networks for large-scale image recognition.
	\newblock {\em CoRR}, abs/1409.1556, 2014.
	
	\bibitem{fashionmnist}
	H.~Xiao, K.~Rasul, and R.~Vollgraf.
	\newblock Fashion-mnist: a novel image dataset for benchmarking machine
	learning algorithms, 2017.
	
	\bibitem{lecun-98}
	Y.~B. Y.~LeCun, L.~Bottou and P.~Haffner.
	\newblock Gradient-based learning applied to document recognition.
	\newblock {\em Proceedings of the IEEE, 86(11):2278-2324, November 1998}.
	
\end{thebibliography}
\end{document}